\documentclass[shortAfour,sageh,times]{sagej}

\usepackage{moreverb,url}

\usepackage[colorlinks,bookmarksopen,bookmarksnumbered,citecolor=red,urlcolor=red]{hyperref}
\usepackage{comment}
\usepackage{amsmath, amssymb, amsfonts}

\usepackage{graphics}
\usepackage[caption=false,font=footnotesize,labelfont=sf,textfont=sf,justification=centering]{subfig}

\usepackage{dblfloatfix}
\usepackage{siunitx}

\newcommand\BibTeX{{\rmfamily B\kern-.05em \textsc{i\kern-.025em b}\kern-.08em
T\kern-.1667em\lower.7ex\hbox{E}\kern-.125emX}}

\def\volumeyear{20XX}

\begin{document}

\runninghead{Aydin et al.}

\title{Towards Collaborative Drilling with a Cobot Using Admittance Controller
}

\author{Yusuf Aydin\affilnum{1*}, Doganay Sirintuna\affilnum{1*}, and Cagatay Basdogan\affilnum{1}}



\begin{abstract}

In the near future, collaborative robots (cobots) are expected to play a vital role in the manufacturing and automation sectors. It is predicted that workers will work side by side in collaboration with cobots to surpass fully automated factories. In this regard, physical human-robot interaction (pHRI) aims to develop natural communication between the partners to bring speed, flexibility, and ergonomics to the execution of complex manufacturing tasks. One challenge in pHRI is to design an optimal interaction controller to balance the limitations introduced by the contradicting nature of transparency and stability requirements. In this paper, a general methodology to design an admittance controller for a pHRI system is developed by considering the stability and transparency objectives. In our approach, collaborative robot constrains the movement of human operator to help with a pHRI task while an augmented reality (AR) interface informs the operator about its phases. To this end, dynamical characterization of the collaborative robot (LBR IIWA 7 R800, KUKA Inc.) is presented first. Then, the stability and transparency analyses for our pHRI task involving collaborative drilling with this robot are reported. A range of allowable parameters for the admittance controller is determined by superimposing the stability and transparency graphs. Finally, three different sets of parameters are selected from the allowable range and the effect of admittance controllers utilizing these parameter sets on the task performance is investigated.

\end{abstract}


\maketitle

\noindent  \affiliation{\affilnum{1}College of Engineering, Koc University, Turkey\\
\affilnum{*}Doganay Sirintuna and Yusuf Aydin contributed equally to this work. They are both co-first authors.\\
Corresponding Author: Yusuf Aydin, Koc University, 34450, Istanbul, Turkey. E-mail: yaydin@ku.edu.tr.}

\pagebreak
\section{Introduction}

The research area that investigates how to integrate human and robot into tasks that involve physical interaction for improvements in performance is known as physical human-robot interaction (pHRI). From assembly tasks to furniture relocation in home/office setting~\citep{Mortl2012}, from industrial applications~\citep{Wojtara2009}, surgery~\citep{Tavakoli2006,Tabatabaei2019} to rehabilitation~\citep{Pehlivan2016,akgun2020}, pHRI may bring high performance solutions to complex problems~\citep[please refer to extensive reviews by][]{Ajoudani2018,MarciaReview}.

To make pHRI more natural, we need robots that can anticipate the intentions of human partner and comply with those intentions smoothly during the execution of a collaborative task. The intention is a state of mind, which cannot be measured directly. However, it is known that humans are good at recognizing each other’s intentions during a collaborative task, even without verbal communication. For instance, while two people transport a table collaboratively, they use the haptic communication channel very effectively. Here, the intended movement can be conveyed to the robot via the direction of the force applied to the table by human partner while the force magnitude helps with the speed of movement. Following the intention recognition, the collaborating human partners successfully adjust their forces to adapt not only to the requirements of task, but also to each other’s needs. Similarly, in pHRI, the robot should not only execute the task, but assist the human partner to perform the task in accordance with her/his intentions. For this purpose, admittance (or impedance) based controllers~\citep{Hogan1984,KeeminkAdmittanceReview2018} have been widely used in pHRI to regulate the interaction between the force applied by human and the velocity of robot's end-effector. In an admittance control architecture, the control parameters must be tuned carefully considering the requirements of collaborative task. These parameters affect both the stability and transparency (i.e., robot’s resistance to human movements) of the human-robot (i.e., coupled) system.

Ensuring stability is necessary for any control system. However, in pHRI, it is the most essential issue since human safety has the highest priority. An unstable robot could cause a severe injury to human. On the other hand, it is quite challenging to preserve
stability in pHRI since an interaction controller is sensitive to changes in environment impedance. For example, when an admittance controller faces a stiff environment, pHRI system may
become unstable. Besides, from the robot’s perspective, human is also a part of this environment, and changes in human arm stiffness may also contribute to the instability. 

Carrying out stability analysis of a pHRI system requires some level of human and environment modeling. To overcome the modeling needs, one can utilize the passivity framework. In this approach,
the human operator and the environment are assumed to act as passive elements, not injecting energy to the closed-loop
system, hence, they do not tend to destabilize the coupled system. If the rest of the control system is
designed to be strictly passive, then the closed-loop system is also passive, ensuring the coupled stability, provided that the system is detectable. It is challenging to design a strictly passive system as discretization, noise, and time delays can make the system active, and active systems can be unstable since they do not dissipate energy \citep{Colgate1997}. \cite{colgateZwidth} showed that there is a theoretical passivity condition for stability, and higher damping and sampling frequency both improve the stability of the interaction. However, utilization of a passivity-based approach results in a conservative controller, which adversely affects the transparency of interaction.

\cite{Dohring2003ThePO} showed that the
conservative effect of the passivity approach can be relaxed up to a certain degree by taking
advantage of modeling. However, uncertainties in modeling can make it difficult to
maintain passivity. As an alternative, \cite{hannafordRyu} proposed passivity observer
(PO) and combined it with the passivity controller (PC). PO measures energy flow in and out of
a system, and if an active behavior is detected, PC acts as an adaptive dissipative element
which absorbs the excessive energy measured by the PO. However, the energy-based
methods require estimation of interaction energy or power which cannot be measured
directly. Moreover, the energy is estimated based on measurements of force and velocity,
which are affected by sampling time, noise, and quantization. Due to the effects as such, an implementation may be challenging. Furthermore, abrupt engagement of PC may disturb the quality of interaction.

Alternatively, knowing that non-passive systems are not necessarily unstable~\citep{Buerger2001}, less conservative controllers can be designed by utilizing linear models of robot, human, and environment, and the effect of modeling parameters on the stability of coupled system can be investigated. \cite{Tsumugiwa2004} carried out root locus analysis for a pHRI task and found that an increase in the environment stiffness is the primary reason for instability, which was also supported by later studies \citep{yusufWHC,Aydin2018,Ferraguti2019}. They argued that when the environment stiffness is high, the system can be stabilized by increasing the damping parameter of the admittance controller. \cite{Duchaine2008} conducted several pHRI
experiments to estimate human arm stiffness offline. They utilized this information to perform stability analysis and adjust the critical damping of the
admittance controller that makes the system stable. However, this method is based on
an offline characterization of maximum human arm stiffness and neglects its time varying character during a pHRI task. On the other hand, the stiffness of human arm depends on its posture, and the amount of activation of the antagonist muscles, which vary with time during the execution of a task \citep{RyuHSO,Lecours}.

Although the exact models of human arm and/or environment impedance cannot be known exactly, the bounds of them can be estimated. For instance,
even though the dynamics of human arm changes over time, human arm impedance varies in a relatively limited range~\citep{Dolan1993,Tsuji1995}. Along this line,~\cite{Buerger2007} proposed a complementary stability approach, which takes advantage of partial knowledge (e.g., bounds) on the models of human and/or environment. Similarly,~\cite{Haddadi2010}
showed that the passivity constraints can be relaxed when the bounds of human arm and environment
impedances are known.

In utilizing the bounds of human arm and/or environment impedances, it is typically assumed that stiffness and mass contributions of these impedances may adversely affect stability. However, recent research shows that a special care should also be given to the damping component~\citep{Willaert2011,Tosun2020}, though it has been assumed so far that increase in damping contributes to stability favorably~\citep{colgate1993achievable,colgateZwidth,Colgate1997,weir2008stability}. In particular, it was shown that high damping may result in non-passive systems under specific conditions, so it may as well be dangerous for stability~\citep{Willaert2011,Tosun2020}.

As mentioned earlier, the conservative limits for robust stability may adversely affect the transparency of a closed-loop system~\citep{Ajoudani2020}. A stable interaction controller can only achieve maximum transparency when it operates close to the stability limit~\citep{Tabatabaei2019}, which may be unsafe due to the possible uncertainties. For example, during a collaborative object transportation task which does not require precision, it is desirable to move the object with minimum human effort. In such a case, the values of the admittance parameters should be small to ensure high transparency, which can also be interpreted as a minimum effort requirement \citep{RyuHSO}. In this way, large displacements can be achieved with small forces. On the other hand, transparency can be secondary in some other tasks. For instance, in tasks requiring precision, such as an object placement task in a confined space, the movements must be made in small increments. In such a case, the admittance controller must utilize larger parameters (i.e., admittance mass and damping). Thus, even if the interaction force increases, the movement takes place in small increments, which is more desirable for precise positioning. As can be seen from the examples above, the need for transparency is directly related to the human intention and task requirements.

\subsection{Contributions}

The requirements for robust stability and perfect transparency are contradictory~\citep{Ajoudani2018}. Ensuring conservative stability adversely affects transparency. On the other hand, working close to the stability limit to improve transparency can lead to instability due to uncertainties in the human arm and environment impedances. Moreover, the need for transparency may change depending on the human intention and task requirements. Considering these points, the long-term objective of our research work is to design an optimal admittance controller that can balance the transparency and stability requirements of a pHRI system to maximize task performance. 

In this paper, we present the initial steps of our design towards achieving this long-term objective. First, the structure of our admittance controller architecture is introduced. In order to analyze the stability and transparency characteristics of this architecture, a dynamical model of cobot is needed. A complete dynamical model of a cobot is nonlinear and difficult to obtain. As an alternative, in this study, we propose an easy-to-implement approach that results in a linearized model and demonstrate its implementation with our cobot (LBR IIWA 7 R800, KUKA Inc.). Being able to obtain a linearized model of cobot is beneficial as more focus can be given to the design process of interaction controller, and powerful analysis methods of linear time-invariant (LTI) systems can be used for this purpose. 

Following the cobot characterization, the methods to examine the stability and transparency of our pHRI system are presented. In our earlier work~\citep{yusufWHC,Aydin2018}, we investigated integer and fractional order admittance controllers and compared their stability characteristics for typical values of human arm impedance. On the other hand, as our objective is to present the steps towards developing an optimal controller in this study, a thorough stability analysis is conducted by considering the combinations of extreme bounds of human arm and environment impedances. Although increase in damping of human arm and/or environment is assumed to improve the stability in the literature, we show that it may destabilize the system under specific conditions and this counter intuitive finding must be accounted for during the design.

In our earlier study~\citep{Aydin2018}, using the concept of impedance matching, we compared the transparency performance of integer and fractional order admittance controllers. On the other hand, in this study, a thorough transparency analysis is conducted for an integer order admittance controller using the concept of parasitic impedance. We show that computational investigation of parasitic impedance is beneficial to determine which parameters are more desirable for higher transparency. 

Following, a range of allowable parameters for the admittance controller is determined by superimposing the results of stability and transparency analyses. Finally, three different sets of parameters are selected from this allowable range by making sure that stability is always maintained and the performance of admittance controllers utilizing these parameter sets is investigated. Specifically, a collaborative drilling task is considered to assess the performance of these controllers.

\section{Control Architecture}

\subsection{Admittance Control}
To regulate a physical interaction between human and robot, impedance/admittance based controllers have been widely used~\citep{Hogan1984}. An impedance controller computes forces to be applied by a robot for an imposed motion, while an admittance controller computes the robot motion for the forces applied by human. In general, it is more natural to utilize admittance control if the robot used for a pHRI task does not posses high backdrivability. Such robots are usually motion controlled; hence, admittance controller generates a motion trajectory based on the interaction force between human and robot, which can be
acquired via sensors attached to the robot. The idea behind this approach is to regulate the
dynamics of interactions between human and robot using a model composed of a mass and
a damper (a spring is not preferred for a collaborative manipulation task since it stores
energy when it is extended/compressed and enforces the robot to return to the
equilibrium/initial position, which is not desirable).

A standard admittance controller has been already used in earlier pHRI studies in the following form:

\begin{equation}
\begin{aligned}
F_\textrm{int} = m\ddot x_\textrm{ref} + b\dot x_\textrm{ref}
\end{aligned}
\end{equation}

\noindent Here, $F_\textrm{int}$ is the interaction force, $m$ and $b$ are the admittance mass and damping, respectively, and $\dot x_\textrm{ref}$, $\ddot x_\textrm{ref}$ are reference velocity and acceleration generated by the admittance controller, respectively. For velocity control, it is practical to write the admittance controller equation in Laplace domain as:

\begin{equation}
\begin{aligned}
V_\textrm{ref}(s) = \frac{F_\textrm{int}(s)}{ms+b}=F_\textrm{int}(s)Y(s)
\end{aligned}
\end{equation}

\noindent Here, $V_\textrm{ref}(s)$ is reference velocity (Laplace transformation of $v_\textrm{ref}=\dot x_\textrm{ref}$), $Y(s)$ is the transfer function of admittance controller.

\subsection{Equivalent Impedance}

\label{Sec:eqImpedance}

\subsubsection{Human Arm Impedance:}

During physical human-robot interaction, human arm impedance changes dynamically. This change might cause admittance controller to be unstable. In stability analyses of pHRI systems, a commonly used simplifying assumption is to model human arm impedance as a second order mass-spring-damper system. Along this line, the impedance of human arm can be modeled as:

\begin{equation}
\begin{aligned}
Z_h(s) = \frac{m_hs^2+b_hs+k}{s}
\end{aligned}
\end{equation}

\noindent Here, $Z_h$ represents the human arm endpoint impedance, $m_h$, $b_h$, and $k_h$ are the effective mass, damping, and stiffness of human arm, respectively.

\subsubsection{Environment Impedance:}

The contact environment can be modelled with a similar linearized model as
\begin{align}
	Z_e(s) = \frac{m_e s^2 + b_e s + k_e}{s}
	\label{Eqn:HumanArmImpedance}
\end{align}
\noindent where $m_e$, $b_e$, and $k_e$ represent mass, damping, and stiffness components of environment model, respectively. 

In our analyses, we combine human arm and environment impedance, and investigate the stability considering the variations in the components of the equivalent impedance. The equivalent impedance used in our study is $Z_{\text{eq}}(s)=\frac{m_{\text{eq}}s^2+b_{\text{eq}}s+k_{\text{eq}}}{s}$, where the equivalent stiffness, damping, and mass are $k_{\text{eq}}=k_h+k_e$, $b_{\text{eq}}=b_h+b_e$, and $m_{\text{eq}}=m_h+m_e$, respectively.

In this study, we consider a pHRI scenario which involves contact interaction with a stiff environment. In particular, we consider a collaborative task of drilling holes on wooden surfaces (i.e., environment) with a drill attached to the end effector of a cobot and manipulated by human operator. In a scenario where a cobot interacts with human but makes no contact with environment, it is sufficient to consider only the limits of human arm impedance. However, in this study, since we consider that cobot interacts with both human and environment, we have evaluated the limits of both human arm and environment impedance together. First, the values required to define human arm impedance were selected in light of the earlier studies in the literature~\citep{Dolan1993,Tsuji1995,Buerger2007}. The selected lower and upper limits were $m_h=0$ and $m_h=5$ kg for the mass, $b_h=0$ and $b_h=41$ Ns/m for the damping and $k_h=0$ and $k_h=401$ N/m for the stiffness of human arm, respectively. In our pHRI scenario, we assume that the environment impedance is represented by a stiffness only, hence, the equivalent damping and mass were kept the same as the damping and mass of human arm, respectively. On the other hand, the equivalent stiffness was assumed to vary in the range of $401 \textrm{ N/m} \leq k_\textrm{eq} \leq 17000\textrm{ N/m}$. Here, the lower limit is determined as the upper bound of human arm stiffness. The upper limit is determined as the total of maximum values of human arm and environment stiffnesses. The maximum stiffness of environment was selected by considering the task of drilling holes on the surface of a workpiece made of strand board \citep{woodChapter}. We varied the equivalent stiffness parameter in this study since the earlier studies suggest that the change in human arm and/or environment stiffness has the greatest impact on the stability of the coupled system~\citep{Tsumugiwa2004,yusufWHC,Aydin2018}.

\subsection{The Closed-Loop System}

\begin{figure}
	\resizebox{\columnwidth}{!}{\rotatebox{0}{\includegraphics{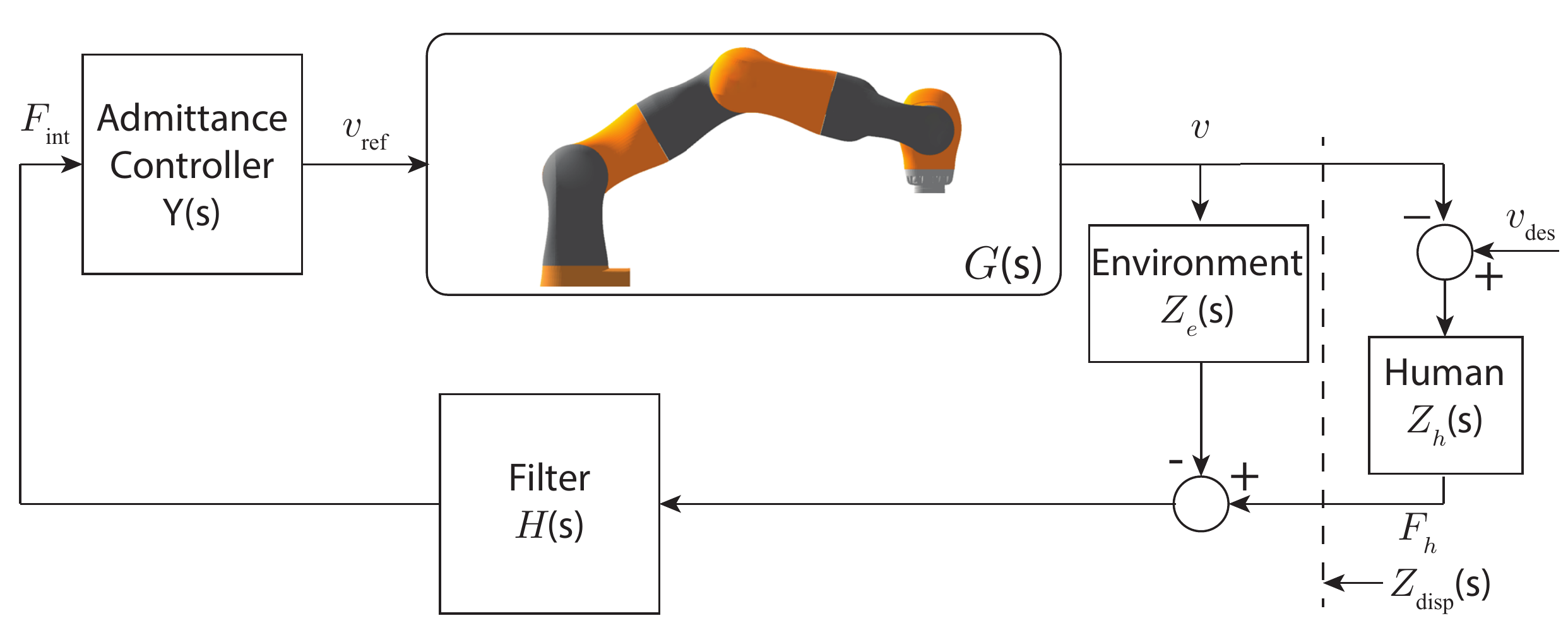}}}
	\caption{Control architecture of our pHRI system\label{Fig:ControlArchitecture}}
	\vspace{-1.5em}
\end{figure}

Figure~\ref{Fig:ControlArchitecture} shows the admittance control architecture used in this study. In this architecture, human and environment impedances are assumed to be coupled ($Z_\textrm{eq}(s) = Z_h(s)+Z_e(s)$), and the resultant interaction force, $F_\textrm{int}$, is measured by the force sensor attached to the end effector of cobot. This measurement is filtered by a low pass filter, $H(s)$, and sent to the admittance controller, $Y(s)$. The controller generates the corresponding reference velocity $v_\textrm{ref}$. Then, cobot's motion controller transmits sufficient torque to its joints in order to achieve this velocity. The cobot's motion controller is robust to the environmental forces acting on its end-effector and it is assumed that the motion of the cobot is not affected by these forces. In Figure~\ref{Fig:ControlArchitecture}, $G(s)$ represents the transfer function of the cobot, and $v$ represents its end-effector velocity.  

The transfer function of the closed loop system is given as:
\begin{align}
	T(s) = \frac{V(s)}{F_\textrm{ref}(s)} = \frac{G(s)Y(s)}{1+G(s)Y(s)H(s)Z_\textrm{eq}(s)}
	\label{Eqn:closedLoopTF}
\end{align}

Using the control architecture given in Figure~\ref{Fig:ControlArchitecture}, we investigate the stability and transparency characteristics of our pHRI system. In our analyses, a second order Butterworth low pass filter having a cutoff frequency of 5 Hz is used for $H(s)$ to account for the sensor dynamics and filtering. However, these analyses also require a model of the cobot, $G(s)$. A dynamical characterization of the cobot to obtain such a model is reported in the next section.

\section{Dynamical Characterization of Cobot}

\begin{figure}
	\resizebox{\columnwidth}{!}{\rotatebox{0}{\includegraphics{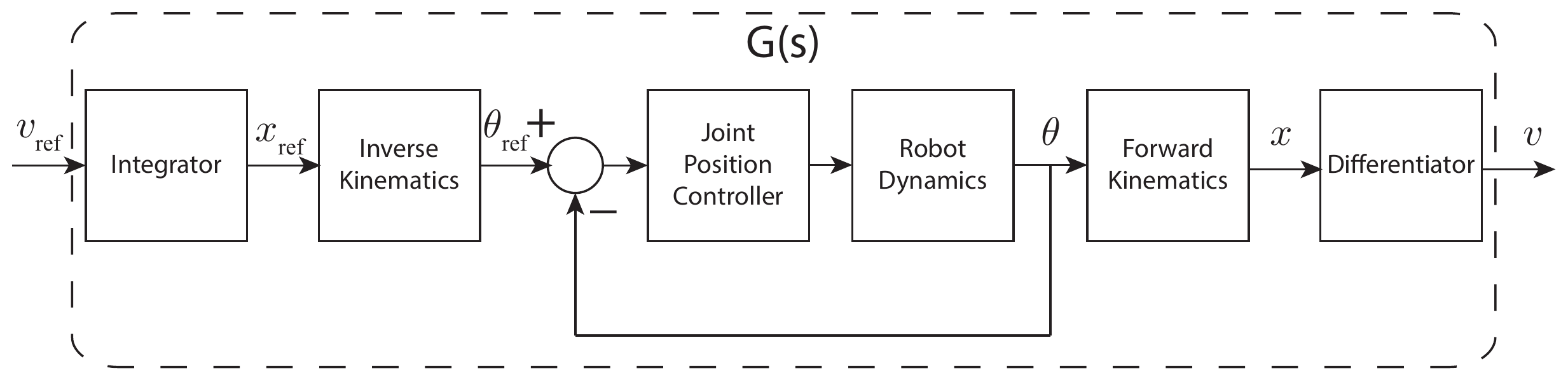}}}
	\caption{Control architecture for LBR iiwa 7 R800 to follow a given $v_\textrm{ref}$ (needed to implement the architecture in Figure~\ref{Fig:ControlArchitecture}). \label{Fig:ControlArchiiwa}}
	\vspace{-1.5em}
\end{figure}

In this study, an LBR iiwa 7 R800 robot (KUKA Inc.) is used for the collaborative drilling task detailed in Section~\ref{Sec:eqImpedance}. Communication with this cobot is possible in two ways: a) via KUKA Sunrise OS system and through Java programming language or b) via Fast Robot Interface (FRI) and through C/C++ programming language. Using the first option, in which the closed-loop is updated at 50 Hz, functions for end effector positioning and point-to-point trajectory tracking are available. The library of the second option (FRI), which is designed for researchers, can reach a high update rate of 1 kHz, but unfortunately does not include such functions. The FRI library contains only the following control functions: a) Joint Position Controller, b) Cartesian Wrench Controller, c) Joint Torque Controller. Since we aim for real-time interaction between human and cobot, the second approach (FRI) is more suitable, but as can be seen from the description above, the FRI library is limited to those 3 control options. To develop an admittance controller for pHRI, we utilized the joint position controller (JPC) of the FRI library. This function enables the cobot to move to the desired joint positions. However, implementation of the admittance controller architecture depicted in Figure~\ref{Fig:ControlArchitecture} requires Cartesian position or velocity controllers, which are not available in the FRI library. As a workaround, the reference velocity $V_\textrm{ref}(s)$ for the cobot was first converted to reference position $X_\textrm{ref}(s)$ by integration and then to the reference joint positions using inverse kinematics (see Figure~\ref{Fig:ControlArchiiwa}). These reference values were then fed to the joint position controller (JPC) which transmits sufficient torque to the corresponding joints in order to follow these position commands. The actual joint positions were obtained from the FRI library, converted to end effector position using forward kinematics and then to velocity by differentiation. The functions for forward and inverse kinematics were adopted from a recent study~\citep{FARIA2018317}.

For the stability and transparency analyses of the control architecture given in Figure~\ref{Fig:ControlArchitecture}, dynamical model of the cobot is needed. Since such a model is not provided by the manufacturer and also not available in the ROS (Robot Operating System) environment, an experimental approach was developed to estimate it along a particular Cartesian axis of motion for a specific joint configuration. Although there are several methods for dynamic characterization of robotic manipulators, they require detailed procedures, and the resulting models are highly nonlinear. For instance, such a complex dynamical model for LBR IIWA 14 R820 (KUKA Inc.), which is the higher payload alternative of our cobot, was recently presented by~\cite{iiwa14}. 

As an alternative, we propose a simple approach to estimate an LTI transfer function model of a cobot, as detailed below. Using this approach, a transfer function between the reference velocity, $V_\textrm{ref}(s)$, and the actual velocity, $V(s)$, of the end effector of our cobot was estimated (Figures~\ref{Fig:ControlArchitecture} and~\ref{Fig:ControlArchiiwa}).

\subsection{Identification Method}

For a certain configuration of the cobot ($\boldsymbol{\theta}_\textrm{nom}$), each joint was commanded to follow a sinusoidal reference joint position profile with a fixed amplitude while the input frequency range was varied from 0.01 to 20 Hz. During these oscillatory movements, the actual joint position was recorded for each joint. Based on the collected data, a transfer function was estimated between the reference and actual positions for each joint of the cobot. 

For each joint $i$, the dynamical relationship between the reference joint position $\theta_{i}^{\textrm{ref}}(s)$ and the actual joint position $\theta_i(s)$ can be defined by a transfer function $T_i(s)$ as following

\begin{equation}
\begin{aligned}
\theta_i(s) &= T_i(s)\theta_i^{\textrm{ref}}(s)
\end{aligned}
\label{Eqn:matchingEqs}
\end{equation}
Basically, the dynamical relationship between $\theta_i^{\textrm{ref}}(s)$ and $\theta_i (s)$ is expected to be the same as the relationship between the reference joint velocity, $\Omega_i^{\textrm{ref}}(s)$, and the actual joint velocity, $\Omega_i(s)$. Hence,

\begin{equation}
\begin{aligned}
\Omega_i(s) &= T_i(s)\Omega_i^{ref}(s)
\end{aligned}
\label{Eqn:matchingEqs}
\end{equation}
is obtained. When all joints are evaluated together, the following equation
\begin{equation}
\begin{aligned}
\boldsymbol{\Omega} (s) &=\boldsymbol{T}(s)\boldsymbol{\Omega}_{\textrm{ref}}(s)
\end{aligned}
\label{Eqn:omega_T_omegaref_relation}
\end{equation}
can be written between the reference joint velocity vector $\boldsymbol{\Omega}_{\textrm{ref}}(s)$ and the actual joint velocity vector $\boldsymbol{\Omega} (s)$. Assuming that the dynamical relationship between $\Omega_i^{\textrm{ref}}(s)$ and $\Omega_i(s)$ for each joint $i$ is independent of the dynamics of the other joints, $T_i(s)$ can be represented by a diagonal matrix

\begin{equation}
\begin{aligned}
\boldsymbol{T} (s) &=\begin{bmatrix} 
    T_{1} (s) & 0  & \dots & 0 \\
    0 & T_2(s) &\dots & 0  \\
    \vdots & \vdots & \ddots & \vdots\\
    0 &   0  & \dots  & T_{7}(s) 
\end{bmatrix}
\end{aligned}
\end{equation}

\noindent When the cobot makes small movements around the $\boldsymbol{\theta}_{\textrm{nom}} $ configuration, the change in the Jacobian matrix of the cobot during these movements can be neglected. In such a case, a fixed $\boldsymbol{J}(\boldsymbol{\theta}_{\textrm{nom}}) $ matrix is obtained using the forward kinematics given by \cite{FARIA2018317}. This Jacobian matrix between the Cartesian velocity vector $\boldsymbol{V}(s)$ and the joint velocity vector $\boldsymbol{\Omega}(s)$ can be written as

\begin{equation}
\begin{aligned}
\boldsymbol{V} (s) &=\boldsymbol{J}(\boldsymbol{\theta}_{\textrm{nom}})\boldsymbol{\Omega}(s) \leftrightarrow \boldsymbol{\Omega}(s)=\boldsymbol{J}^{\dagger}(\boldsymbol{\theta}_{\textrm{nom}})\boldsymbol{V}(s)
\end{aligned}
\label{Eqn:veljacob}
\end{equation}

\noindent where $\boldsymbol{J}^{\dagger}(\boldsymbol{\theta}_{\textrm{nom}})$ is the pseudoinverse of $\boldsymbol{J}(\boldsymbol{\theta}_{\textrm{nom}})$. Likewise, the following equation holds between $\boldsymbol{V}_{\textrm{ref}}(s)$ and $\boldsymbol{\Omega}_{\textrm{ref}}(s)$
\begin{equation}
\begin{aligned}
\boldsymbol{V}_{\textrm{ref}} (s) &=\boldsymbol{J}(\boldsymbol{\theta}_{\textrm{nom}})\boldsymbol{\Omega}_{\textrm{ref}}(s) \leftrightarrow \boldsymbol{\Omega}_{\textrm{ref}}(s)=\boldsymbol{J}^{\dagger}(\boldsymbol{\theta}_{\textrm{nom}})\boldsymbol{V}_{\textrm{ref}}(s)
\end{aligned}
\label{Eqn:velrefjacob}
\end{equation}

\noindent Combining (\ref{Eqn:veljacob}) and (\ref{Eqn:velrefjacob}) and rearranging using (\ref{Eqn:omega_T_omegaref_relation}), the following relationship is obtained between $\boldsymbol{V}_{\textrm{ref}}(s)$ and $\boldsymbol{V}(s)$ 

\begin{equation}
\begin{aligned}
\boldsymbol{V} (s) &=\boldsymbol{K} (s) \boldsymbol{V}_{\textrm{ref}}(s)
\end{aligned}
\label{Eqn:VKVref}
\end{equation}

\noindent where $\boldsymbol{K}(s) = \boldsymbol{J}(\boldsymbol{\theta}_{\textrm{nom}})\boldsymbol{T}(s)\boldsymbol{J^{\dagger}}(\boldsymbol{\theta}_{\textrm{nom}})$.

\subsection{Implementation for Our System}

\begin{figure}
    \centering
    \resizebox{0.4\columnwidth}{!}{\rotatebox{0}{\includegraphics[trim=0cm 0cm 0cm 0cm, clip=true]{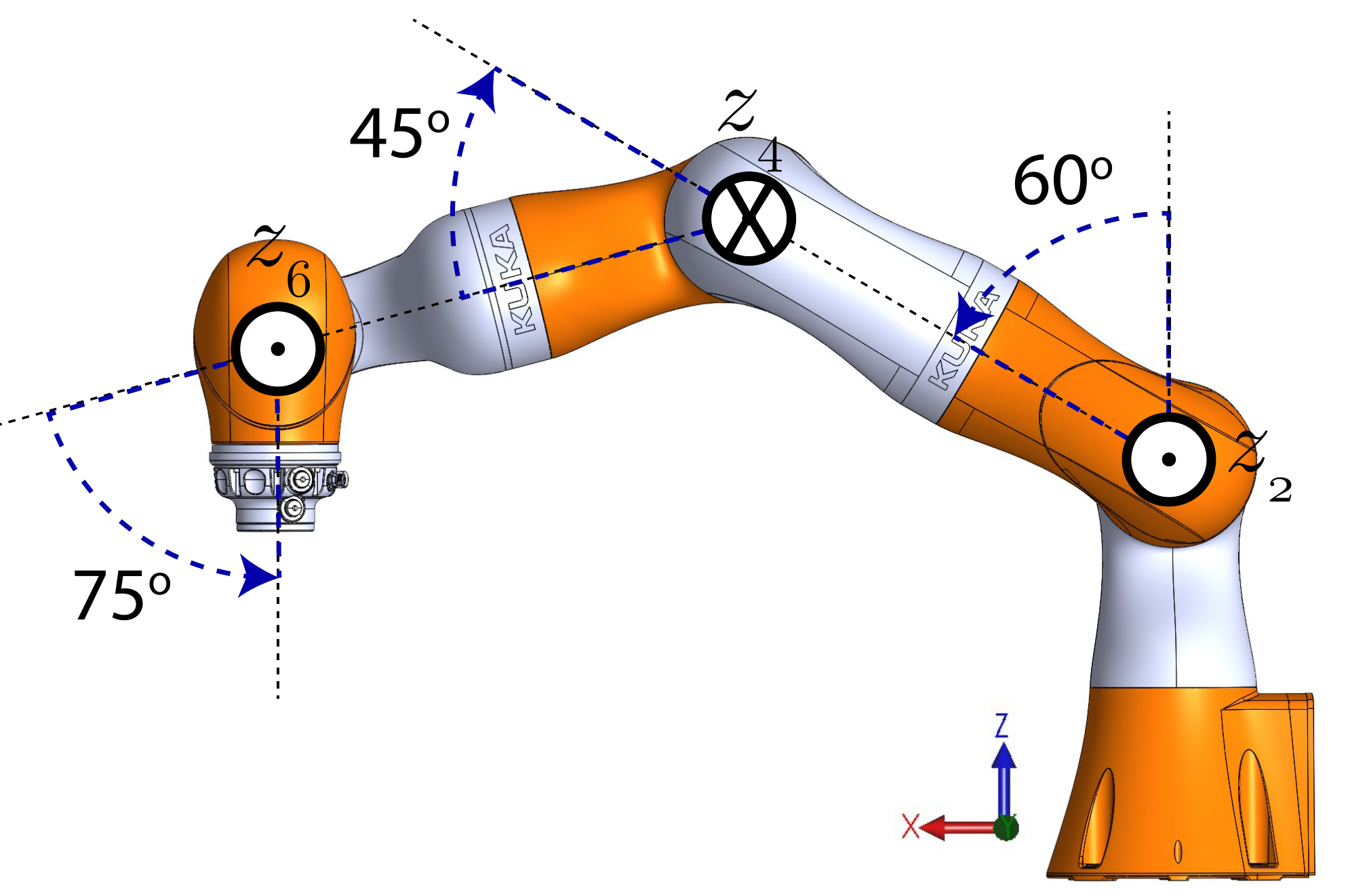}}}
    \caption{The nominal configuration ($\theta_\textrm{nom}$) selected for the dynamical characterization of our robot. Axes $z_2$, $z_4$, and $z_6$ represent the rotation axes of joints 2, 4, and 6, respectively.}
    \label{fig:charConf}
\end{figure}

In our study, the end effector of our cobot is constrained to move along the x-axis direction only with a reference velocity defined in Cartesian space, $v_{x}^{\textrm{ref}}$, while the robot is in $\boldsymbol{\theta}_{\textrm{nom}}$ configuration (Figure~\ref{fig:charConf}). In this case, 
$\boldsymbol{V}_{\textrm{ref}}(s) $ can be defined as

\begin{equation}
\begin{aligned}
\boldsymbol{V}_{\textrm{ref}} (s) &=\begin{bmatrix}
V_{x}^{\textrm{ref}}(s)&0&0&0&0&0 
\end{bmatrix}^T
\end{aligned}
\label{Eqn:velrefVectorInLaplace}
\end{equation}

\noindent where $V_{x}^\textrm{ref}(s)$ is the representation of $v_{x}^\textrm{ref}$ in the Laplace domain. When (\ref{Eqn:velrefVectorInLaplace}) plugged into (\ref{Eqn:VKVref}), the actual velocity vector of the end effector of our cobot can be written in Cartesian space as

\begin{equation}
\begin{aligned}
\boldsymbol{V}(s) &=\begin{bmatrix}
V_{x}(s)  \\
\dots \\
\dots \\
\dots \\
\dots \\
\dots \\
\end{bmatrix}
= 
\underbrace{ 
\begin{bmatrix} 
    K_{11} (s) &  \dots & K_{16}(s) \\
    \vdots & \ddots & \vdots  \\
    K_{61}(s) &    \dots    & K_{66}(s) 
\end{bmatrix}}_{\boldsymbol{K}(s)}
\begin{bmatrix}
V_{x}^{\textrm{ref}}(s)  \\
0 \\
0 \\
0 \\
0 \\
0 \\
\end{bmatrix}
\end{aligned}
\label{Eqn:matchingEqs}
\end{equation}

\noindent where $V_x(s)$ is the Laplace transform of $v_x$, the actual velocity of the end effector along the x-axis direction. Using this equation, the following relation can be obtained between $V_x (s)$ and $V_x^{\textrm{ref}}(s)$ as;

\begin{equation}
\begin{aligned}
\frac{V_{x}(s)}{V_{x}^{\textrm{ref}}(s)} = K_{11}(s)
\end{aligned}
\label{Eqn:matchingEqs}
\end{equation}

\noindent For the dynamic characterization of our cobot along the x-axis, the nominal configuration was selected as (Figure~\ref{fig:charConf}), 
\begin{equation}
\begin{aligned}
\boldsymbol{\theta}_{\textrm{nom}} &=\begin{bmatrix}
0  & \pi/3 & 0 & -\pi/4 & 0 & 5\pi/12 & 0
\end{bmatrix}^T
\end{aligned}
\label{Eqn:charConf}
\end{equation}

\noindent In this configuration, only the joints 2, 4 and 6 are actuated when the end effector of the cobot moves along the x-axis only. Moreover, $K_{11}(s)$ is a linear combination of $T_2(s)$, $T_4(s)$ and $T_6(s)$ only, as expected. In this case, the dynamical model of the cobot, $G(s) = K_{11}(s)$, can be calculated as follows (note that $G(s)$ will be used instead of $K_{11}(s)$ from this point on to be consistent with the notation used for the control architecture given in Figure~\ref{Fig:ControlArchitecture}).

\begin{equation}
\begin{aligned}
G(s) = K_{11}(s)=k_2T_2(s)+k_4T_4(s)+k_6T_6(s),
\\
k_2=-0.1896,\textrm{ } k_4=1.6550,\textrm{ } k_6=-0.4654
\label{Eqn:G=kT}
\end{aligned}
\end{equation}

\begin{figure*}[t]

	\centering
	\begin{minipage}{\textwidth}
		\centering

		\subfloat[$2^\textrm{nd}$ joint\label{fig:j2}]{%
			\resizebox{0.325\columnwidth}{!}{\rotatebox{0}{\includegraphics[trim=0cm 0.0cm 0.0cm 0.0cm, clip=true]{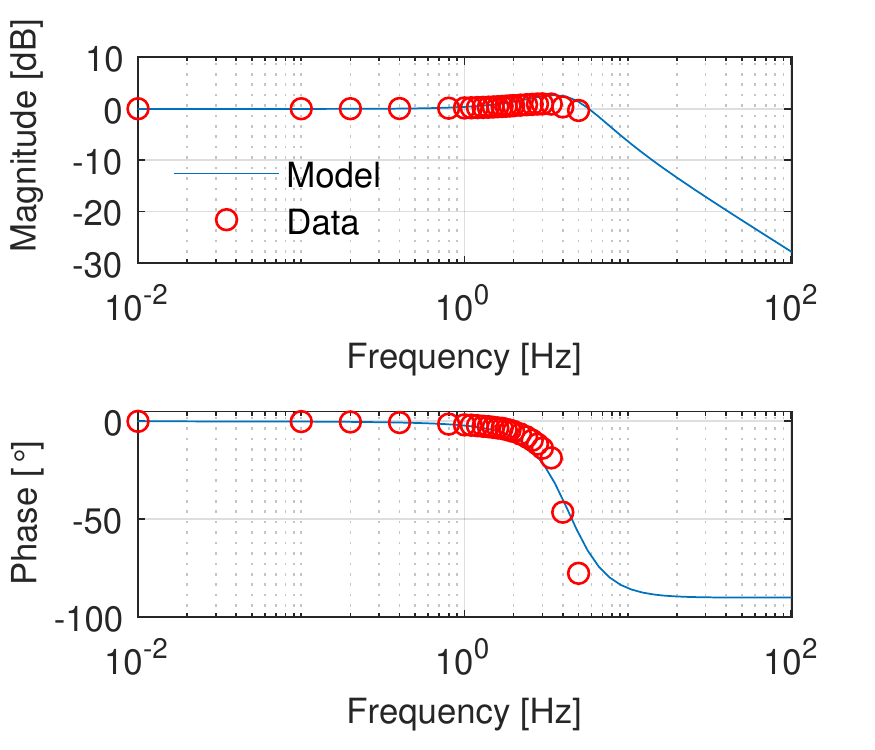}}}
		}
		\hfill
		\subfloat[$4^\textrm{th}$ joint\label{fig:j4}]{%
			\resizebox{0.325\columnwidth}{!}{\rotatebox{0}{\includegraphics[trim=0cm 0.0cm 0.0cm 0.0cm, clip=true]{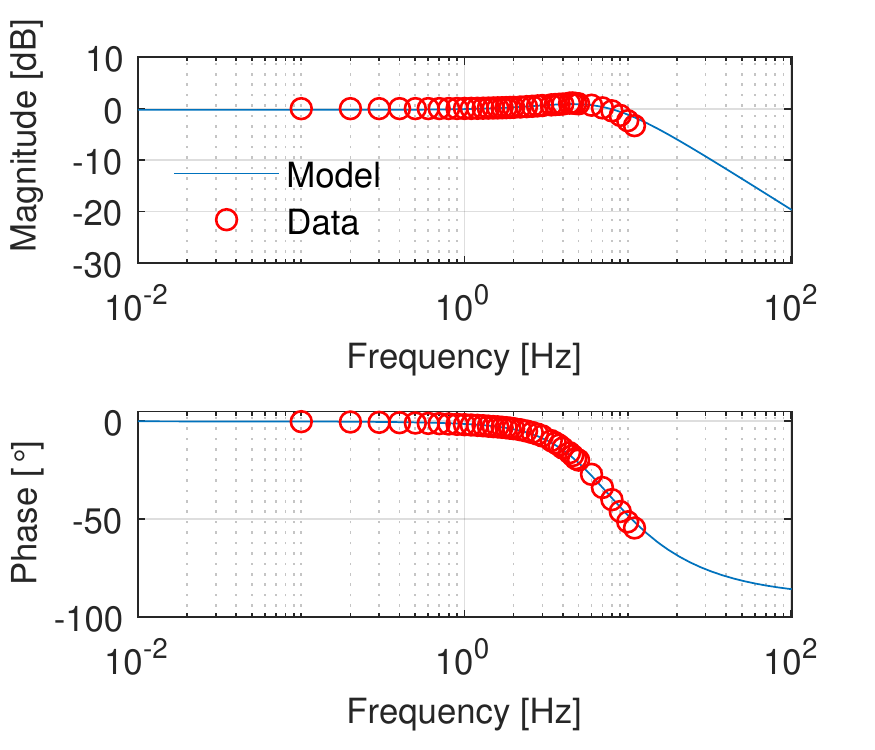}}}
		}
				\hfill
		\subfloat[$6^\textrm{th}$ joint\label{fig:j6}]{%
			\resizebox{0.325\columnwidth}{!}{\rotatebox{0}{\includegraphics[trim=0cm 0.0cm 0.0cm 0.0cm, clip=true]{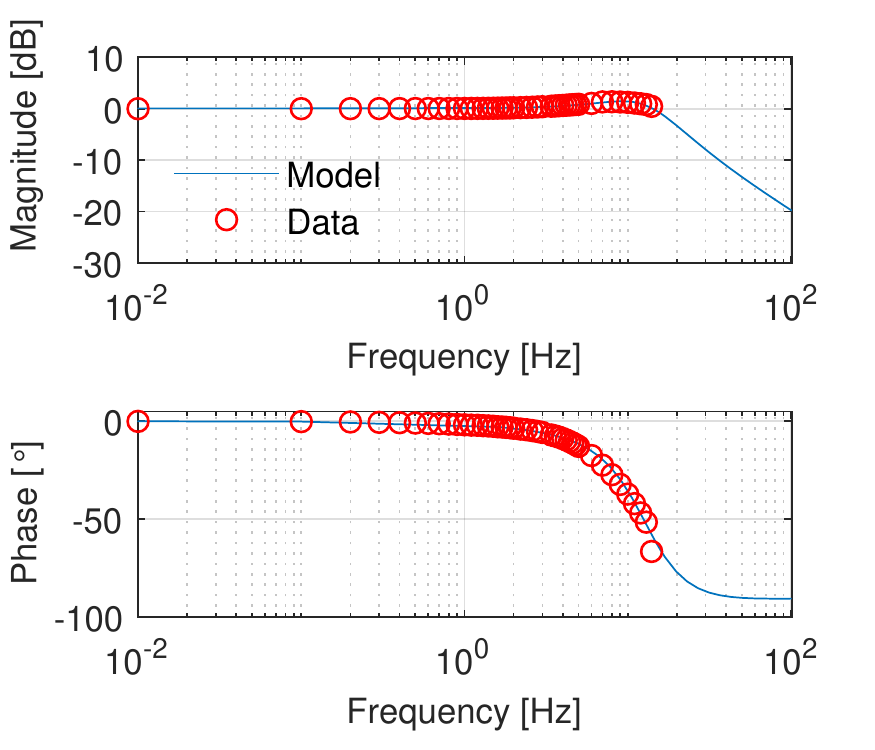}}}
		}
	\end{minipage}

	\caption{Magnitude and phase angle as a function of frequency for joints a) 2, b) 4, and c) 6.}
	\label{Fig:charData}
\end{figure*}

As can be seen from the equation above, it is sufficient to perform dynamical characterization of those three joints of the cobot only (2, 4, and 6). For this purpose, sinusoidal joint position commands were sent to those joints of the cobot to oscillate them individually for 180 seconds at frequencies varying from 0.01 to 20 Hz. This frequency range was selected by considering the fact that natural human arm movement is band-limited~\citep[typically $\leq2$ Hz,][]{Brooks1990,Dimeas2016}. The duration of 180 seconds for oscillations was determined by considering the time at which the transients in dynamical response of each joint vanished, and continuous oscillations at the commanded frequency was reached. The actual joint position values during the oscillations were recorded using the internal sensors of the cobot. All data were collected at 1000 Hz, the highest frequency allowed by the FRI interface. Actual (output) and reference (input) joint position signals were examined in the frequency domain, and the ratio of the amplitudes of input and output signals and the phase difference between them were calculated for each frequency. Using this information, a transfer function, $T_i(s)$, was estimated for each of those joints (2, 4, 6). First and second order polynomials were chosen for the numerator and denominator of this transfer function, respectively 
\citep[assuming that the motion of each joint of the cobot is controlled by a PD controller, the numerator of the transfer function was estimated as a first order polynomial, while the denominator was estimated as a second order polynomial considering the motor and arm dynamics, in light of][]{craig}. Thus, for the aforementioned joints, the following transfer functions were obtained between the actual and reference joint positions utilizing the adaptive subspace Gauss-Newton search method available in the System Identification Toolbox of MATLAB software (MathWorks Inc.).

\begin{equation}
\begin{aligned}
T_{2}(s)&=\frac{\theta_{2}(s)}{\theta_{2}^{\textrm{ref}}(s)}&=\frac{25.69s +749.3}{s^2 + 29.04s +752.7}\\
T_{4}(s)&=\frac{\theta_{4}(s)}{\theta_{4}^{\textrm{ref}}(s)}&=\frac{65.99s +1679}{s^2 + 72.97s +1723}\\
T_{6}(s)&=\frac{\theta_{6}(s)}{\theta_{6}^{\textrm{ref}}(s)}&=\frac{63.77s +6564}{s^2 + 95.39s +6513}
\end{aligned}
\label{Eqn:jointTFs}
\end{equation}

For the aforementioned joints of the cobot, the magnitude and phase angle values obtained from the experimental data, and
the ones returned by the transfer function models are shown in Figure~\ref{Fig:charData}. As shown in Figure~\ref{fig:j2}, the frequency bandwidth of joint 2 is the lowest since it carries all the remaining links and motors. On the other hand, the bandwidth of joint 6 is the highest (see Figure~\ref{fig:j6}) as it carries the least load. Moreover, for any given frequency, the phase delay is the least at joint 6 whereas it is the highest at joint 2.

\begin{figure}
    \centering
    \subfloat[\label{fig:rlocus}]{
    \resizebox{0.4\columnwidth}{!}{\rotatebox{0}{\includegraphics[trim=0cm 0cm 0cm 0cm, clip=true]{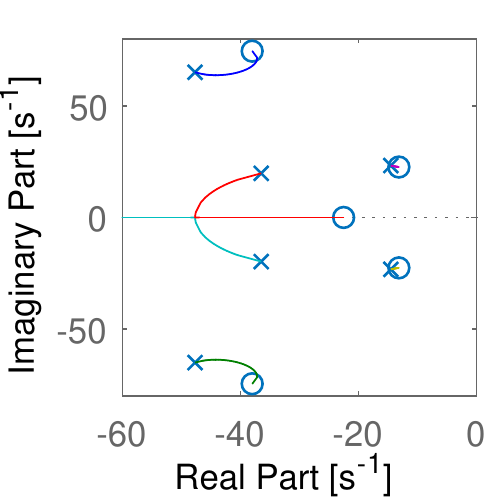}}}
    }
    \subfloat[\label{fig:bode}]{
    \resizebox{0.43\columnwidth}{!}{\rotatebox{0}{\includegraphics[trim=0cm 0cm 0cm 0cm, clip=true]{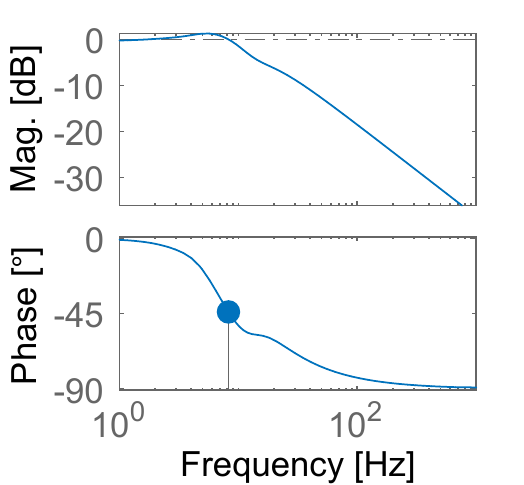}}}
    }
    \caption{a) Root locus of the robot transfer function, $G(s)$. b) Bode magnitude and phase of the robot transfer function, $G(s)$ (the distance of the solid blue dot to -180 degree is 136 degrees, which is the phase margin.)}
    \label{fig:rlocBode}
\end{figure}

Using (\ref{Eqn:G=kT}) and the transfer functions given in (\ref{Eqn:jointTFs}), obtaining the overall transfer function of the robot $G(s)$ is trivial. Since $T_2(s)$, $T_4(s)$ and $T_6(s)$ are stable, we expect that $G(s)$ is also stable since it is a linear combination of those. When we examine the root-locus of $G(s)$, we observe that all the roots are in the left half plane (Figure~\ref{fig:rlocus}). Figure~\ref{fig:bode} shows the Bode magnitude and phase angle of the estimated transfer function of the cobot. Note that the stability of the dynamic model is sufficiently robust as the phase margin is 136 degrees. 

In the following sections, we report the stability and transparency analyses of the coupled system obtained by using the estimated dynamical model of the cobot.

\section{Stability Analysis}

\label{stability}

\begin{figure}[t]

	\centering

		\centering

			\resizebox{0.5\columnwidth}{!}{\rotatebox{0}{\includegraphics[trim=0.05cm 0.05cm 0.45cm 0.2cm, clip=true]{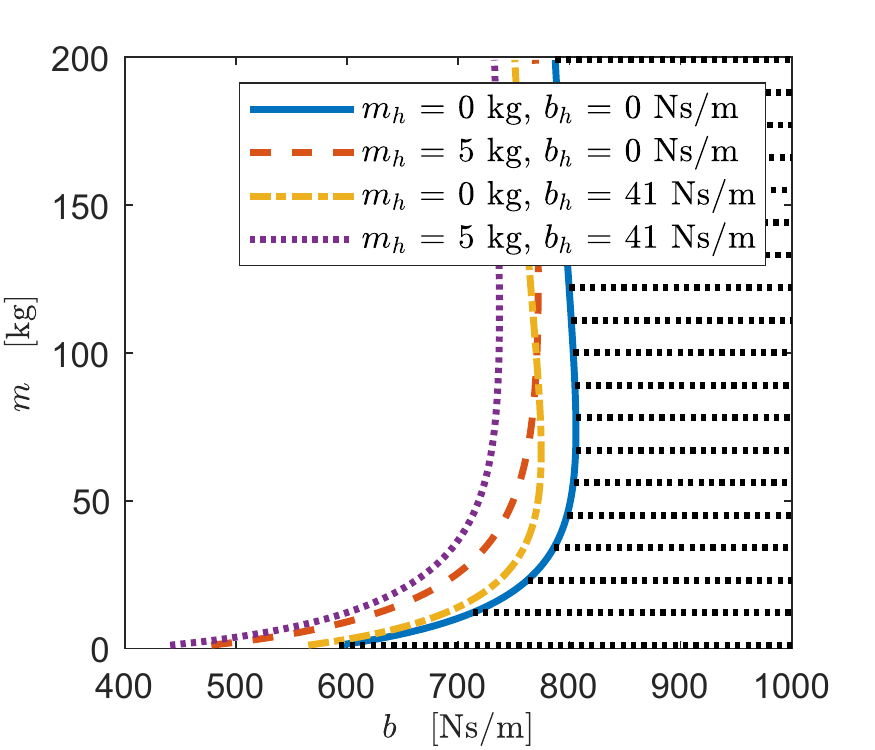}}}

	\caption{Stability map of our pHRI system for $k_{\text{eq}}=17000$ N/m. The curves represent the stability boundaries for different combinations of extreme values of $m_{\text{eq}}$, and $b_{\text{eq}}$. The shaded region shows the sets of admittance parameters that ensure stability for all combinations of extreme bounds of $m_{\text{eq}}$, and $b_{\text{eq}}$.}
	\label{Fig:stabMaps}
\end{figure}

Considering the boundaries of equivalent impedance, the stability characteristics of the coupled system is analyzed using stability maps. A stability map is a graphical representation of the controller parameters for which the resulting closed-loop system becomes stable for ranges of environment and human arm impedances. 
The stability analysis for a pHRI system can be performed by numerically studying the location of closed-loop poles. This requires some knowledge of the dynamics of the human arm and the environment, which is not easy to obtain due to uncertainties in these. Although the exact dynamics of human arm and environment are uncertain, their behavior typically resides within specific limits. For this reason, a computational approach for designing a pHRI controller should focus on the boundaries of the parameter space of the equivalent impedance.

In Figure~\ref{Fig:stabMaps}, the stability map of our pHRI system is depicted. This map was generated by considering the bounds of the equivalent impedance for the pHRI scenario given in Section~\ref{Sec:eqImpedance}. It is a valid assumption that mass and damping parameters of human arm impedance may not change drastically during the execution of a pHRI task. However, it is difficult to know their exact values in practice. Moreover, in case human releases her/his contact with the end-effector of cobot at any instant during the interaction, the mass and damping contribution of human arm reduces to zero which may destabilize the system. So, in our controller design process, for a practical implementation, the bounds for $m_h$ and $b_h$ were considered for the bounds of $m_{\text{eq}}$ and $b_{\text{eq}}$, respectively.

As shown in Figure~\ref{Fig:stabMaps}, as $m$ increases, it is necessary to use higher $b$ to stabilize the system. On the other hand, for a given $m$, the minimum $b$ required to stabilize the system increases with the reduction of $m_{\text{eq}}$. In other words, as $m_h$ increases, it is more difficult to stabilize the system under low $m$. Given that the exact value of the equivalent mass $m_{\text{eq}}$ is uncertain (due to the lack of information on $m_h$), any reduction in the mass contribution of human arm $m_h$ can easily destabilize the system. For example, suppose the controller parameters are selected as $m = 50$ kg and $b = 780$ Ns/m, taking into account only the upper limit of $m_{\text{eq}}$. If a person completely releases her/his contact with the cobot's end effector, the stable region shrinks and this currently selected set of controller parameters shifts to the unstable zone. Thus, the stability of the coupled system is jeopardized. To prevent such an undesirable possibility from occurring, a designer should consider these effects.

\begin{figure}[t]

	\centering

		\centering

			\resizebox{0.5\columnwidth}{!}{\rotatebox{0}{\includegraphics[trim=0.05cm 0.05cm 0.45cm 0.2cm, clip=true]{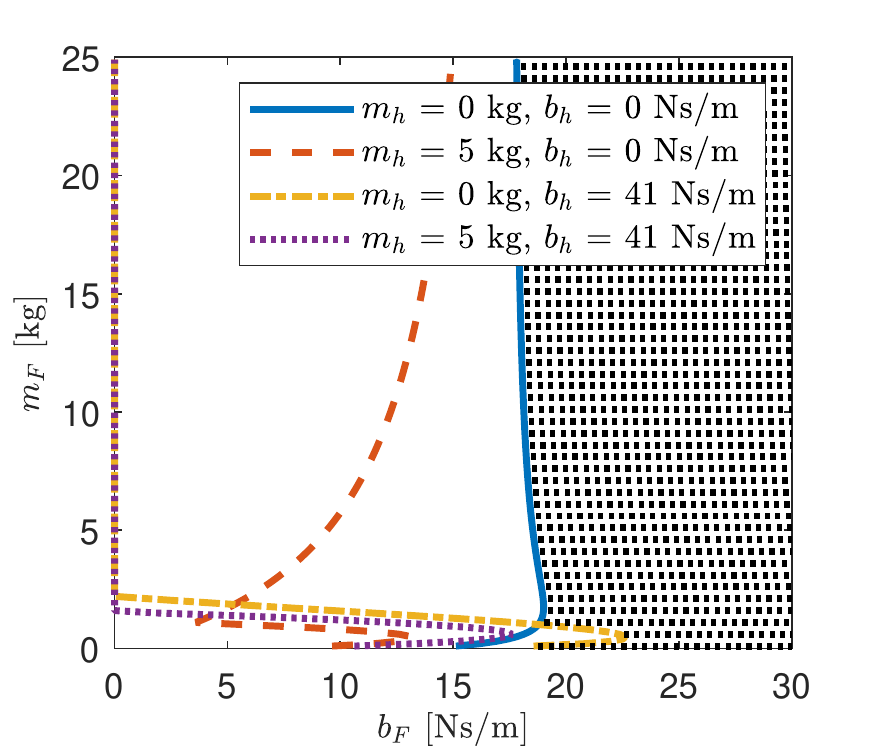}}}

	\caption{Exemplary stability map for $k_e=0$ ($k_{\text{eq}}=k_h=401$ N/m). The curves represent the stability boundaries for different combinations of extreme values of $m_{\text{eq}}$, and $b_{\text{eq}}$. The shaded region show the sets of admittance parameters that ensure stability for all combinations of extreme bounds of $m_{\text{eq}}$, and $b_{\text{eq}}$.}
	\label{Fig:stabMap0kh}
\end{figure}

Moreover, it is widely accepted that $b_{\text{eq}}$ contributes to the stability favorably~\citep{colgate1993achievable,colgateZwidth,Colgate1997, weir2008stability}, thus, only the lower bound of $b_{\text{eq}} = 0$, is considered as the worst case in the earlier studies.
However,~\cite{Willaert2011} and \cite{Tosun2020} presented several examples where larger $b_{\text{eq}}$ may result in an active (i.e., non-passive) system under specific conditions, though the system is passive for $b_{\text{eq}}=0$. Instead of passivity, we directly investigated the stability in our analysis, and observed that our results do not indicate such a situation within the limits of the equivalent impedance considered. However, if the limits of equivalent impedance are different from those studied here, further analysis is needed to check whether stability is compromised with increased physical damping as passivity is. Along this line, we considered another task where $k_e=0$, and generated the corresponding stability map depicted in Figure~\ref{Fig:stabMap0kh}. For this task, let’s assume that admittance controller parameters are set to $m = 0.5$ kg and $b = 17$ Ns/m considering only the lower bound of the equivalent damping ($b_\text{eq}$ = 0) as the worst case, in line with the assumption that $b_\text{eq}$ contributes to the stability favorably~\citep{colgate1993achievable,colgateZwidth,Colgate1997, weir2008stability}. However, an increase in the physical damping easily destabilizes the system since the lowest allowable value for $b$ is 23 Ns/m when the equivalent damping in the system increases to its maximum value (Figure~\ref{Fig:stabMap0kh}). Therefore, the upper bound of the physical damping in the equivalent impedance should be considered in addition to its lower bound to guarantee stability.

Along these lines, a controller parameter set which makes the system stable under all combinations of extreme bounds of $m_{\text{eq}}$ and $b_{\text{eq}}$ was considered as a stable set, so that any change in $m_{\text{eq}}$ and $b_{\text{eq}}$ could be tolerated. In light of this, shaded area  in Figure~\ref{Fig:stabMaps} was accepted as the stable region for the pHRI scenario given in Section~\ref{Sec:eqImpedance}.

\section{Transparency}
\label{Sec:Transparency}

The impedance that human feels during interactions
with an environment has critical importance in pHRI tasks~\citep{colonnese2016,OzanSpringer}. For
instance, in a robotic-assisted surgery, the control architecture
for interaction should ensure that the impedance of soft tissue is reflected to the surgeon so that she/he can have a
realistic understanding of the surgical operation, which is not shadowed
by the robot’s dynamics. Alternatively, in a different task,
human may wish to move robot in free space, where she/he
does not desire to feel any dynamics due to the robot. In a drilling task, one can argue that the former example is more suitable (i.e., reflecting the environmental impedance to human without any parasitic effects of cobot). In other words, human can guide cobot to advance a drill into a workpiece by appreciating its stiffness and the dynamics of interactions. Along this line, the closed-loop impedance displayed to human is investigated in this section.

The closed-loop impedance displayed to human operator, $Z_{\text{disp}}(s)$,
(see Figure~\ref{Fig:ControlArchitecture}) can be written
as:
\begin{align}
    Z_{\text{disp}}(s) = \frac{F_h(s)}{V(s)} = \frac{1+G(s)Y(s)H(s)Z_{e}(s)}{G(s)Y(s)H(s)}
    \label{Eqn:closedLoopImpedance}
\end{align}

\noindent The parasitic impedance $\Delta Z(s)$ is defined as the difference between the
desired impedance $Z_{\text{des}}(s)$ and the impedance reflected to human
$Z_{\text{disp}}(s)$;
\begin{align}
    \Delta Z(s) \triangleq Z_{\text{des}}(s)-Z_{\text{disp}}(s)
    \label{Eqn:diffImp}
\end{align}

\noindent As long as the parasitic impedance $\Delta Z(s)$ is small (large), then the
transparency of the coupled system is high (low). In our pHRI application, the
desired impedance equals to the environment impedance,
$Z_{\text{des}}(s)=Z_e(s)$. Using this information, together with
\eqref{Eqn:closedLoopImpedance} and \eqref{Eqn:diffImp}, the magnitude of
parasitic impedance in frequency domain is given as
\begin{align}
    |\Delta Z(j\omega)| = 1/|G(j\omega)Y(j\omega)H(j\omega)|
    \label{Eqn:delZSimple}
\end{align}

\noindent Clearly, maximizing $|{GYH}| $ minimizes $|\Delta Z|$, which in turn
maximizes the transparency. Furthermore, we can deduce that maximizing the
magnitude of $Y(j\omega)$ is required to improve the transparency. For this purpose, minimizing the magnitude of its denominator, $|m(j\omega) + b|$ is sufficient.

As expected, lower values of $m$ and $b$ result in higher transparency.
Especially at lower frequencies, the effect of $b$ is more dominant on the
parasitic impedance. Therefore, lower values of $b$ are more desirable for
higher transparency at low frequencies. Moreover, the effect of $m$
becomes more dominant at higher frequencies, and lower values of $m$ are also
more desirable for higher transparency at high frequencies.

\begin{figure}[t]

	\centering

		\centering

			\resizebox{0.48\columnwidth}{!}{\rotatebox{0}{\includegraphics[trim=0cm 0.05cm 0.68cm 0.2cm, clip=true]{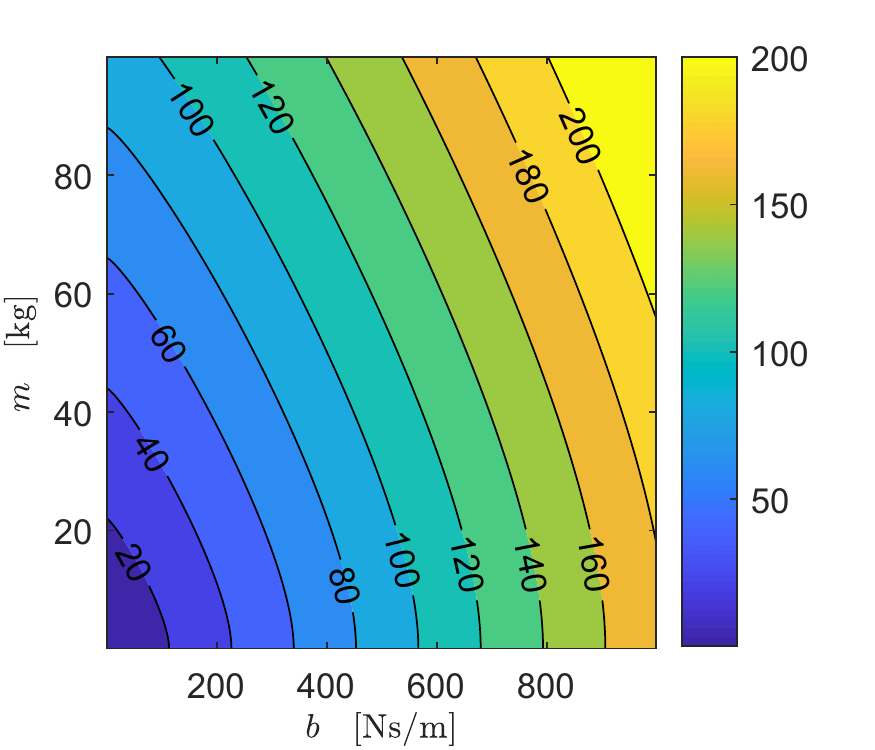}}}

	\caption{Parasitic impedance cost map of our pHRI system.}
	\label{Fig:parMaps}
\end{figure}

Inspecting the magnitude of $Y(j\omega)$ alone gives information about how controller parameters affect the transparency. However, for more conclusive results, parasitic impedance function should be inspected since the cobot itself contributes to the parasitic impedance as well. The fact that the contributions of controller and cobot on parasitic impedance change as functions of frequency calls for a quantative metric for parasitic impedance. A proper one was defined by~\cite{Buerger2007} which computes it for a discrete range of frequencies. Adopting that cost function, the following measure for parasitic impedance cost is defined

\begin{align}
C = \sum_{\omega_L}^{\omega_U}W(\omega)\log |\Delta Z(j\omega)|
\label{Eqn:parasiticcost}
\end{align}

\noindent where $W(\omega)$ is a weight function which can be used to adjust the contribution of each frequency. In addition, $\omega_L$ and $\omega_U$ are the lower and upper boundaries of the frequency range. In this study, a logarithmically spaced frequency range varying from 0.01 to 30 Hz was chosen for the discretization of the parasitic impedance function. The magnitude of a fifth order Butterworth filter having a cutoff frequency of 5 Hz was used as a weight function to boost the effect of low frequency content on the parasitic impedance cost and reduce the contribution of higher frequencies as the frequency range of human voluntary movements is limited~\citep{Brooks1990,Dimeas2016}. Thus, achieving lower parasitic impedance for higher transparency is more desirable at lower frequencies. Using (\ref{Eqn:parasiticcost}), parasitic impedance cost was evaluated for each set of controller parameters, and the resulting parasitic impedance cost map was constructed (Figure~\ref{Fig:parMaps}). As expected, lowering $m$, and $b$ reduce the parasitic impedance cost, leading to higher transparency.

\begin{figure}[t]

	\centering

		\centering

			\resizebox{0.48\columnwidth}{!}{\rotatebox{0}{\includegraphics[trim=0.05cm 0.05cm 0.45cm 0.2cm, clip=true]{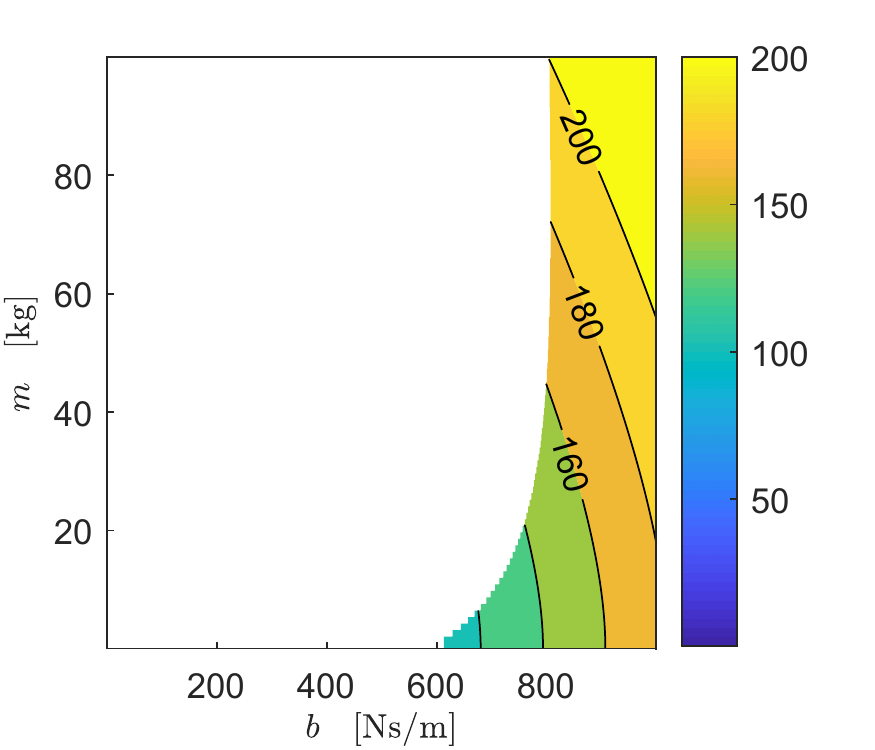}}}

	\caption{Map illustrating the allowable controller parameters, obtained by combining the stability (Figure~\ref{Fig:stabMaps}) and parasitic impedance cost maps (Figure~\ref{Fig:parMaps}). The controller parameters in white region are not allowed. The colored region depicts allowable parameters with the corresponding parasitic impedance cost.}
	\label{Fig:allowPars}
\end{figure}

The value of the parasitic impedance cost, so as the level of transparency, is not influenced by the environment impedance for a given set of controller parameters, though using lower controller gains to achieve higher transparency may not be feasible due to the stability constraints (see Figure~\ref{Fig:stabMaps}). Therefore, stability and transparency should be considered together. Along this line, allowable controller parameters that combine stability and transparency objectives are given in Figure~\ref{Fig:allowPars}. As can be seen, transparency degrades (i.e., the cost of parasitic impedance increases) as $b$ increases, while the stability of pHRI system improves.

\vspace{-0.5\baselineskip}

\section{Experimental Evaluation}
Task performance under three different controllers, whose parameters were selected from the allowable range given in Figure~\ref{Fig:allowPars}, were compared for our pHRI task. In this task, an experimenter (an expert user of the system) manipulated a power drill that was rigidly attached to the end effector of our cobot to open a hole on a wooden workpiece made of strand board (see Figure~\ref{fig:expsetup}).

\subsection{Controller Parameters}

We selected three conservative sets of admittance parameters which can maintain the stability robustly at the contact interface (see Table 1).

\begin{table}[h]
\small\sf\centering
\caption{Controller parameters selected for the drilling task.}
\begin{tabular}{ccc}
\toprule
 & m {[}kg{]} & b {[}Ns/m{]} \\
\midrule
I & 20 & 1500 \\
II & 20 & 900 \\
III & 50 & 900\\
\bottomrule
\end{tabular}
\end{table}

\subsection{Experimental Setup}
\begin{figure}
    \centering
    \resizebox{0.6\columnwidth}{!}{\rotatebox{0}{\includegraphics[trim=0cm 0cm 0cm 0cm, clip=true]{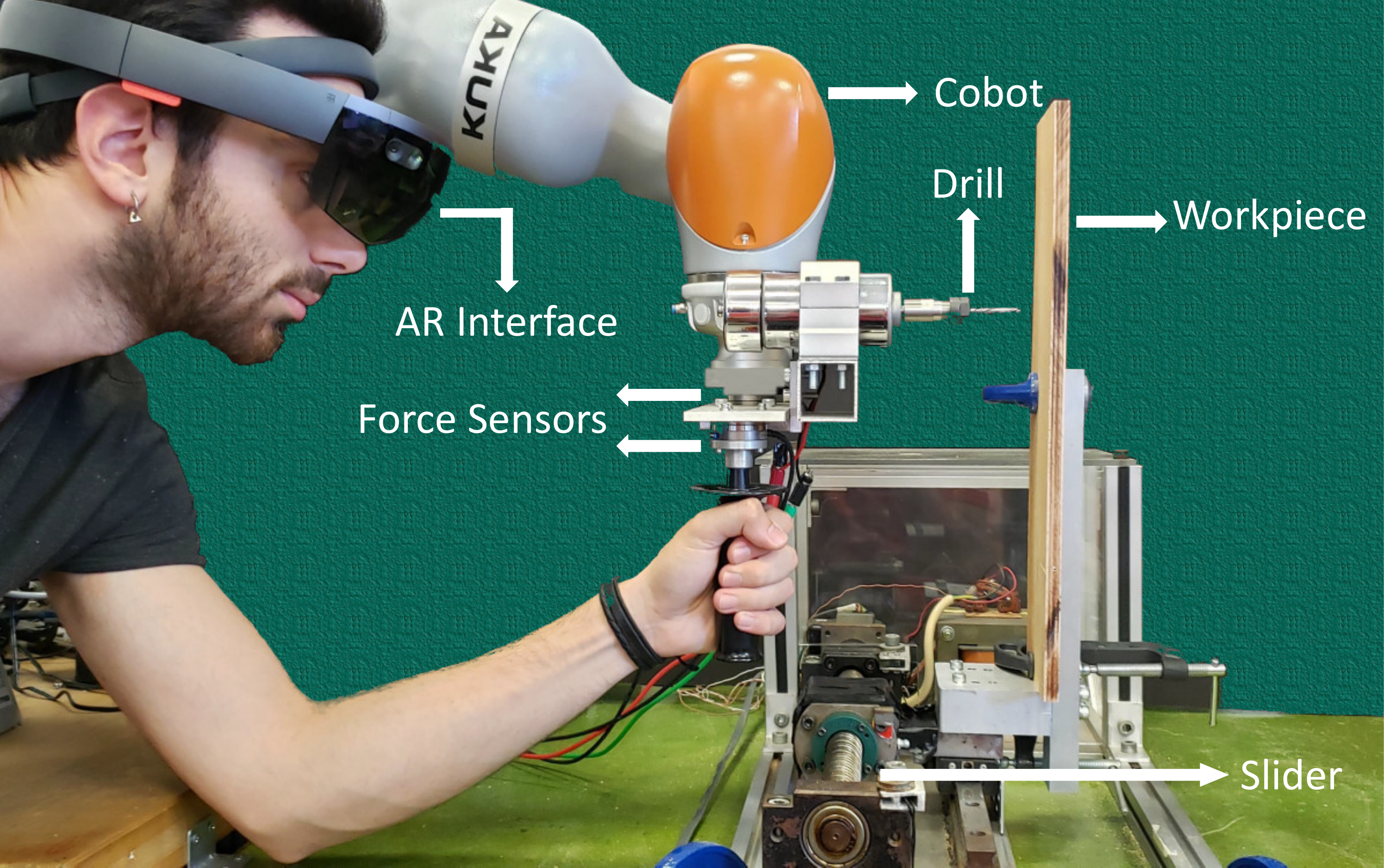}}}
    \caption{The components of our physical human-robot interaction system for collaborative drilling. The cobot constrains the movement of user to help with the task while the AR interface informs about its phases.}
    \label{fig:expsetup}
\end{figure}

The major components of our experimental setup are a power drill, two force sensors (Mini40, ATI Inc.), and a handle attached to the end effector of LBR iiwa 7 R800 robot (KUKA Inc.) as shown in Figure~\ref{fig:expsetup}. One of the force sensors was used to measure the interaction force between the drill bit and the workpiece, while the other one measured the force applied by the user alone. The force data from both sensors were acquired simultaneously at \SI{10}{\kilo \hertz} using a DAQ card (PCI-6225, National Instruments Inc.), though the control loop given in Figure~\ref{Fig:ControlArchitecture} was updated at \SI{500}{\hertz}. In order to reach this high update rate, we utilized the Fast Robot Interface (FRI) library of the cobot and implemented the admittance controller by ourselves in C/C++ using the “Joint Position Controller” function of the library and by taking advantage of the forward and inverse kinematics reported in~\cite{FARIA2018317}. The workpiece was a flat wooden block made of strand board with dimensions of \SI[product-units = single]{400 x 110 x 10}{\milli\meter}, which was placed on a moving stage at a distance of $d = \SI{95}{\milli\meter}$ from the tip of drill bit. Upon completion of the task, the stage was commanded to move the workpiece in tangential direction, aligning the next drilling location and the direction of cobot movement. The AR interface used in this study (HoloLens, Microsoft Inc.) informed the user about the distance of the drill tip to the workpiece (see Figure~\ref{fig:drive}), and instantaneous drilling depth with respect to the targeted drill depth of \SI{5}{\milli\meter} (see Figure~\ref{fig:drill}).

\subsection{Experimental Procedure}
\begin{figure}[ht!]
    \centering
    \subfloat[Driving phase]{
         \includegraphics[width=0.248\textwidth]{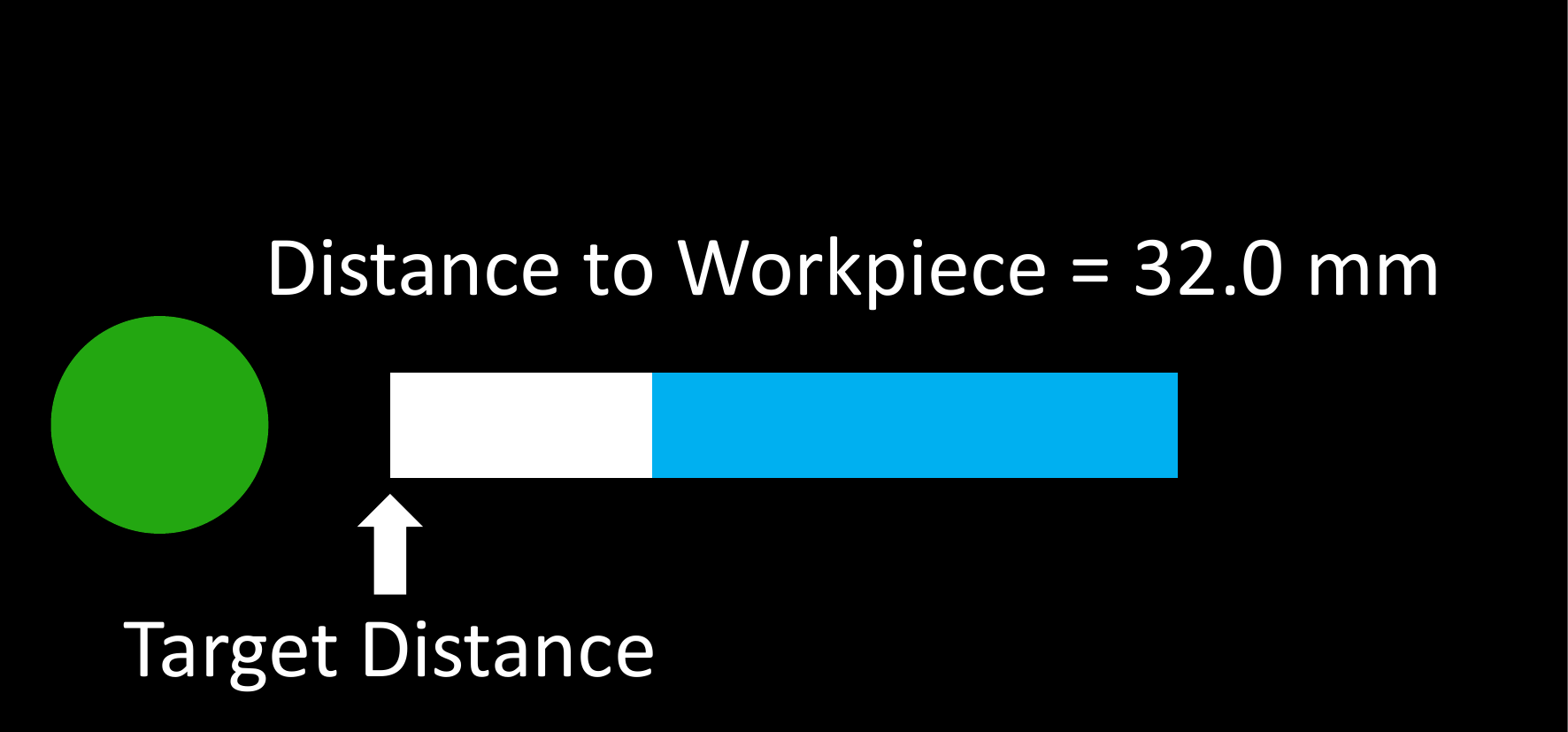}\label{fig:drive}}
    \subfloat[Penetration phase]{
         \includegraphics[width=0.24\textwidth]{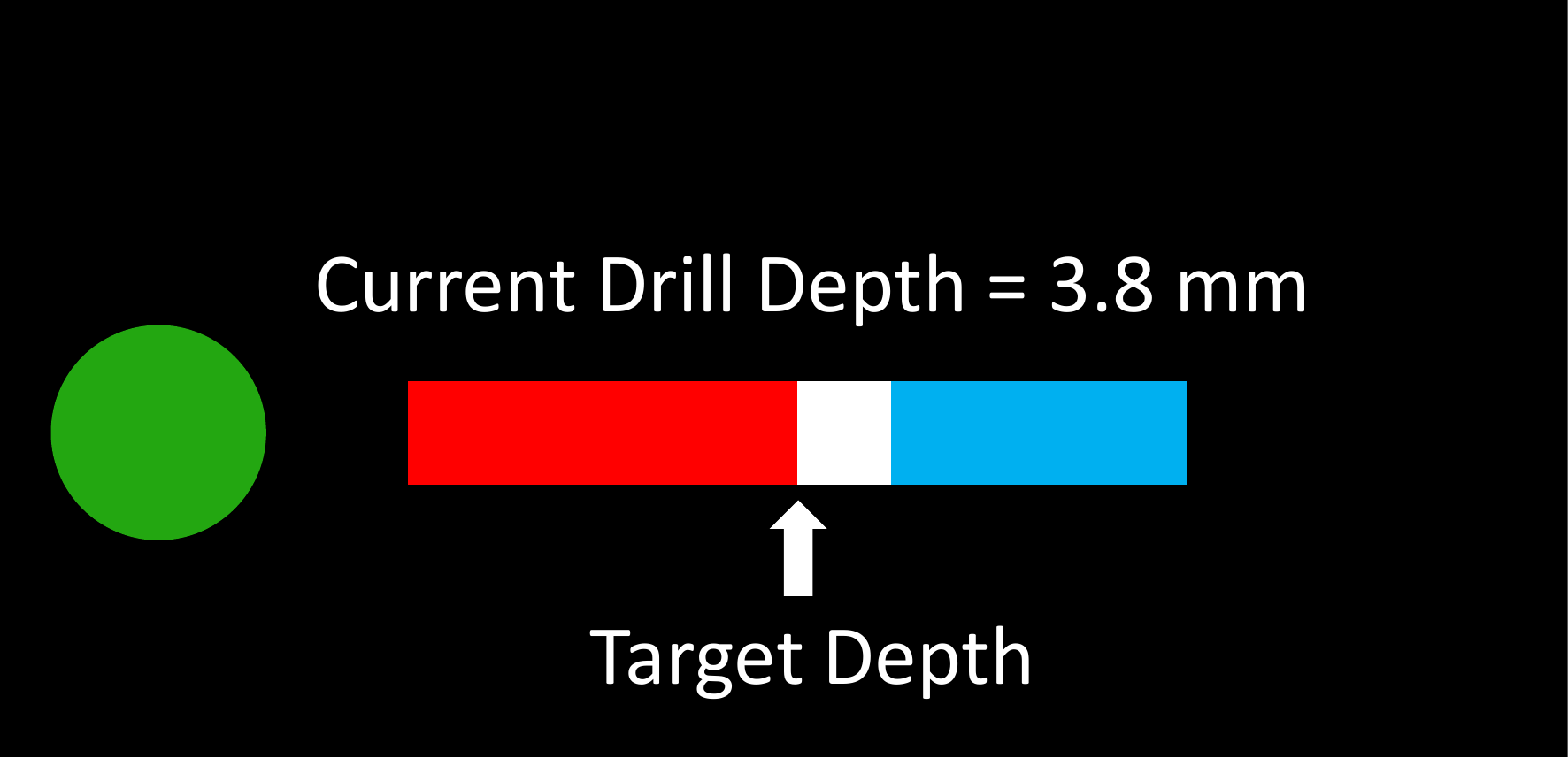} \label{fig:drill}}
    \caption{Visual information displayed to the experimenter through the AR interface during the execution of collaborative drilling task.}
    \label{fig:bars}
\end{figure}

The experimenter performed a total of \num{12} trials (3 interaction controllers $\times$ 4 repetitions). In all trials, the cobot constrained the movements of the experimenter along the direction perpendicular to the surface of workpiece (i.e., x-axis). Each trial consisted of 1) a driving phase, where the experimenter gripped the handle with his dominant hand and guided the cobot towards the drilling location (see Figure~\ref{fig:drive}), and 2) a penetration phase, which started with the initial contact of the drill bit to the workpiece and ended when the desired depth was reached. In the penetration phase, the experimenter penetrated into the workpiece to open a hole with a depth of \SI{5}{\milli\meter}; the current drill depth was displayed through the visor of AR interface (see Figure~\ref{fig:drill}). In the driving phase, distance to workpiece was displayed through the visor, whereas the current drill depth was displayed during the penetration phase only. In Figure~\ref{fig:bars}, the green circle indicates that the controller is active. The red and white bars represent the cross-section of the wooden workpiece. The red bar shows the part that should not be penetrated, whereas the white bar shows the desired depth that will be drilled. The blue bar grows from right to left as the drill bit penetrates into the workpiece, and shows the current drill depth.

\subsection{Results and Discussion}

\begin{figure}
    \centering
    \resizebox{0.6\columnwidth}{!}{\rotatebox{0}{\includegraphics[trim=0cm 0cm 0cm  0.2cm, clip=true]{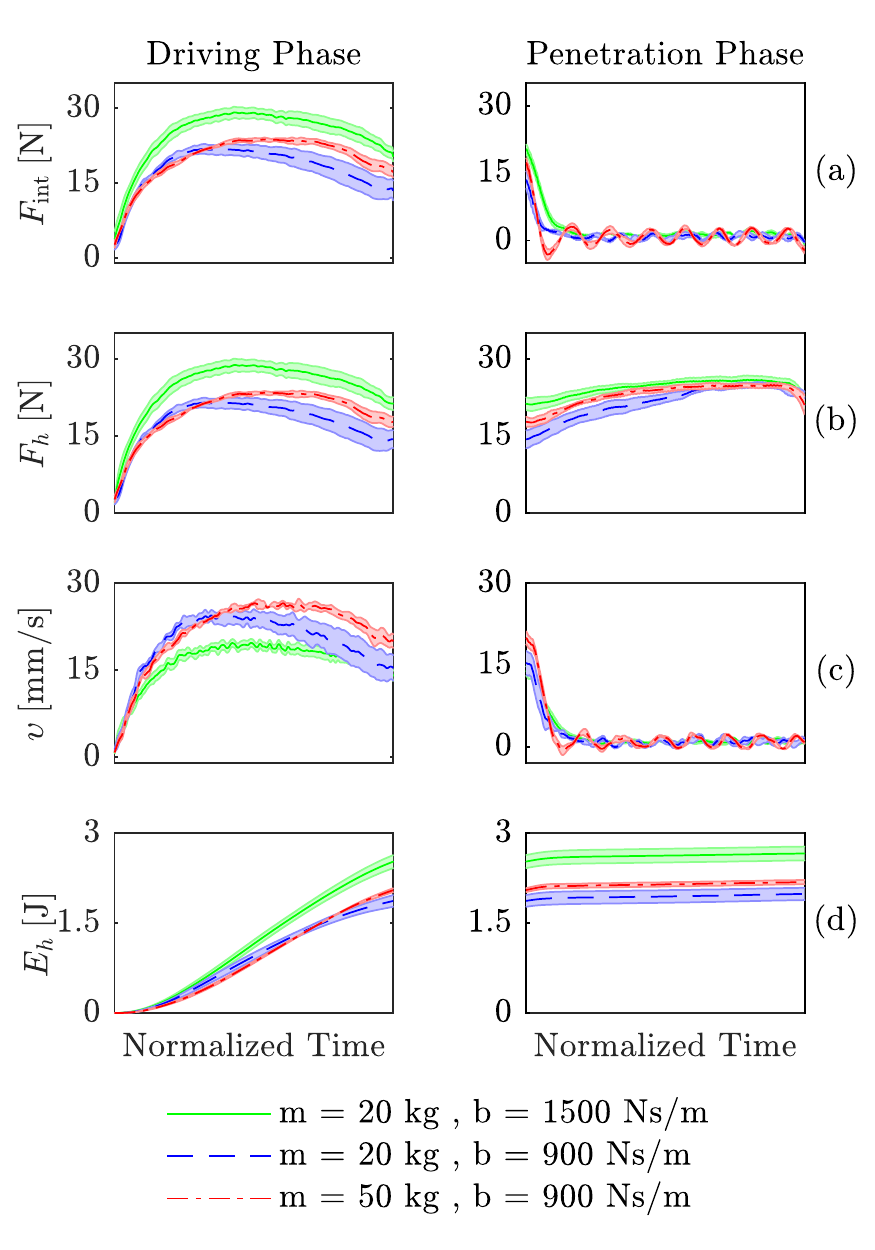}}}
    \caption{The interaction force, $F_\textrm{int}$, the force applied by the experimenter, $F_h$, the velocity of the end effector, $v$, and the effort made by the experimenter, $E_h$, as functions of normalized time. The light shaded regions represent the standard deviations of 4 repetitions.}
    \label{fig:timeplots}
\end{figure}

\begin{figure}
    \centering
    \vspace{1.2em}
    \resizebox{0.6\columnwidth}{!}{\rotatebox{0}{\includegraphics[trim=0cm 0cm 0cm 0cm, clip=true]{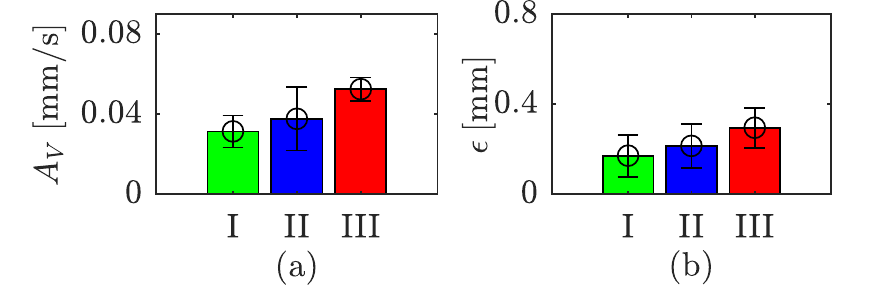}}}
    \caption{The means and standard errors of means for the peak amplitude of oscillations for the end-effector velocity and the absolute value of error in drilling depth (I: $m = 20$ kg, $b= 1500$ Ns/m , II: $m = 20$ kg, $b= 900$ Ns/m, III: $m = 50$ kg, $b= 900$ Ns/m).}
    \label{fig:effortbar}
\end{figure}

Interaction force, $F_\textrm{int}$, force applied by the experimenter, $F_h$, velocity of the end effector, $v$, and effort made by the experimenter, $E_h$, under each controller are depicted in Figure~\ref{fig:timeplots} as functions of normalized time for both phases. Moreover, the mean value of peak amplitude of oscillations for the end-effector velocity $A_{V}$ and the absolute value of error in drilling depth $\epsilon$ are illustrated for the penetration phase in Figure~\ref{fig:effortbar}.

\newcommand{\RomanNumeralCaps}[1]
    {\MakeUppercase{\romannumeral #1}}

Examining Figure~\ref{fig:timeplots}, we observed that the interaction force, $F_\textrm{int}$, the force applied by the experimenter, $F_h$, and the effort made by the experimenter, $E_h$, were lower at low admittance damping ($b = 900$ Ns/m) during the driving phase, while the velocity of the end effector was higher. Therefore, the coupled system (human-robot-environment) was more transparent under low damping, as expected. Since the movement took place at a low frequency during the driving phase, the change in the admittance mass was not expected to affect the performance of the pHRI system significantly (Figure~\ref{fig:timeplots}). Such an effect was expected to occur during contact interactions with environment (e.g., the penetration phase in our study). As one can observe, the amplitude of the oscillations during the penetration phase was higher under higher admittance mass (Figures~\ref{fig:timeplots}a,~\ref{fig:timeplots}c, and~\ref{fig:effortbar}a). Increasing the mass while keeping the damping constant moved the system closer to stability boundary (see Figures~\ref{Fig:stabMaps}, and~\ref{Fig:allowPars}), hence the dissipation capacity of the controller became less effective to damp out the end-effector oscillations sufficiently. Furthermore, such an action reduced the robustness of the pHRI system. In addition, utilizing higher admittance mass increased the absolute error in drilling depth (Figure~\ref{fig:effortbar}b), as it was not easy to accelerate and decelerate the system under such high admittance mass.

\section{Discussion and Conclusion}

In this paper, our initial steps towards developing an optimal admittance controller that can balance the stability and transparency requirements of a pHRI system was introduced. First, a dynamical model of our cobot was developed in order to analyze the characteristics of the closed-loop system. In particular, the approach utilized to characterize our cobot was a simple and practical one as it can be implemented to any robotic manipulator with ease. Furthermore, although the proposed approach was implemented for a single Cartesian direction in this study (i.e., x-axis) considering the requirements of the pHRI task (i.e., collaborative drilling), it is general enough to be utilized to obtain a dynamical model of a cobot for all directions in Cartesian space. Moreover, as the resulting model of cobot is going to be a linearized one, powerful tools of LTI system design can be easily used for analyzing a closed-loop pHRI system incorporating this model.

Following the dynamical characterization of our cobot, the stability and transparency analyses of the coupled system were conducted and the results were presented in the form of graphical maps. In particular, the stability map provided the allowable range for the controller parameters, and the parasitic impedance map depicted the corresponding transparency characteristics. In our earlier work~\citep{yusufWHC,Aydin2018}, the stability analysis was carried out for typical values of human arm impedance in order to compare admittance controllers of two different forms. In this study, to account for the possible uncertainties in human arm and environment impedances, the stability analysis was carried out by considering the bounds of the equivalent impedance (i.e., all extreme combinations of the parameters of human arm and environment impedances). Although it is typically assumed that higher values of equivalent damping contribute to the stability favorably~\citep{colgate1993achievable,colgateZwidth,Colgate1997, weir2008stability}, we illustrated an example that an increase in the physical damping can destabilize the system, if not accounted for. Initially, we found no evidence for instability under high equivalent damping within the limits of the equivalent impedance considered in our pHRI task (Figure~\ref{Fig:stabMaps}). However, we showed that if the impedance of environment varies (e.g., workpieces having different impedance characteristics are to be drilled), higher values of equivalent damping can destabilize the system (Figure~\ref{Fig:stabMap0kh}). Similarly,~\cite{Willaert2011} and \cite{Tosun2020} reported several examples where having higher environment damping could result in an active system (i.e., non-passive). Although~\cite{Willaert2011} and \cite{Tosun2020} showed that higher physical damping may yield in a non-passive system under some special conditions, we know that such systems are not necessarily unstable~\citep{Buerger2001}. To this end, our result complements the findings by~\cite{Willaert2011} and \cite{Tosun2020}, and suggests that the upper bound of the physical damping should be considered for the stability analysis in addition to its lower bound.

The transparency analysis shows that lower values of $m$ and $b$ result in higher transparency. Especially at lower frequencies, the effect of $b$ is more dominant on the parasitic impedance. Considering that the frequency range of the intentional movements of human arm is band-limited~\citep[$\leq$ 2 Hz,][]{Brooks1990,Dimeas2016}, lower damping is more desirable for higher transparency. This is also supported by the parasitic impedance map depicted in Figure~\ref{Fig:parMaps}.

Finally, the task performance under admittance controllers with three different sets of parameters were experimentally evaluated and compared. In particular, the controllers were tested for a collaborative drilling task, which is a practical application demonstrating the advantages of collaborative execution of a pHRI task. Once the drill bit is advanced into a workpiece along its surface normal, any positional deviations of it in other directions can easily break the drill bit and cause harm to the relevant hardware and human user. Considering this, the cobot in our study constrained the movements of the user along the drill direction during the task. Such a constraint also allowed us to test the controllers in a rigorous manner and present the results in a more digestible form.

In our future studies, we plan to investigate stability robustness and develop a metric to quantify it as we have done it for the transparency in this study by defining a metric for parasitic impedance. Following, we aim to combine the metrics for stability robustness and parasitic impedance and develop a general framework to design an optimal admittance controller which can balance the transparency and stability requirements of a pHRI system to maximize task performance.

\begin{acks}
The authors thank to Prof. V. Patoglu, Dr. O. Tokatli, and Dr. O. Caldiran for fruitful discussions made during the course of this study. The Scientific and Technological
Research Council of Turkey (TUBITAK) supported this work
under contract EEEAG-117E645.

\end{acks}

\bibliographystyle{SageH}
\bibliography{references}

\end{document}